\documentclass[conference]{IEEEtran}
\IEEEoverridecommandlockouts
\usepackage{amsmath,amssymb,amsfonts}
\usepackage{algorithmic}
\usepackage{graphicx}
\usepackage{textcomp}
\usepackage{mathtools}
\usepackage{commath}
\usepackage{subcaption}
\usepackage[english]{babel}
\usepackage{siunitx}    
\usepackage{xcolor}
\usepackage[ruled,vlined,linesnumbered]{algorithm2e}
\usepackage{cite}
\usepackage{float}
\usepackage[utf8]{inputenc}
\usepackage[T1]{fontenc} 

\usepackage{fancyhdr,lipsum}
\fancypagestyle{firstpage}{% Page style for first page
\fancyhead[C]{To appear at the 23rd Design, Automation and Test in Europe (DATE 2020) in Grenoble, France.}% Header
\fancyfoot[C]{\thepage}% Footer
}

\begin{document}
\font\myfont=cmr12 at 22pt
\font\alfont=cmr12 at 10pt

\title{\myfont FT-ClipAct: Resilience Analysis of Deep Neural Networks and Improving their Fault Tolerance\\ using Clipped Activation}

\author{\IEEEauthorblockN{Le-Ha Hoang, Muhammad Abdullah Hanif, Muhammad Shafique }
\IEEEauthorblockA{\textit{Technische Universität Wien (TU Wien), Vienna, Austria} \\
                         {\{le-ha.hoang, muhammad.hanif, muhammad.shafique\}@tuwien.ac.at}                       
                }        
        }        
\maketitle

\thispagestyle{firstpage}

\begin{abstract}
%Deep Neural Networks (DNNs) are widely being adopted for safety-critical applications, e.g., healthcare and autonomous driving. Inherently, they are considered to be highly error-tolerant. However, recent studies have shown that hardware faults that impact the parameters of a DNN (e.g., weights) can have drastic impacts on its classification accuracy.  In this paper, we perform a comprehensive error resilience analysis of DNNs subjected to hardware faults (e.g., permanent faults) in the weight memory. The outcome of this analysis is leveraged to propose a novel error mitigation technique which squashes the high-intensity faulty activation values to alleviate their impact. We achieve this by replacing the unbounded activation functions with their clipped versions. We also present a method to define the clipping values of the activation functions that result in increased resilience of the networks against faults. We evaluate our technique on the modified AlexNet network, using CIFAR-10 dataset. The experimental results show that our mitigation technique significantly improves the network's resilience to faults. For example, the proposed technique offers on-average 25\% improvement in the classification accuracy of the modified AlexNet model at \num{5e-7} fault rate, when compared to the network without any fault mitigation.
Deep Neural Networks (DNNs) are widely being adopted for safety-critical applications, e.g., healthcare and autonomous driving. Inherently, they are considered to be highly error-tolerant. However, recent studies have shown that hardware faults that impact the parameters of a DNN (e.g., weights) can have drastic impacts on its classification accuracy. In this paper, we perform a comprehensive error resilience analysis of DNNs subjected to hardware faults (e.g., permanent faults) in the weight memory. The outcome of this analysis is leveraged to propose a novel error mitigation technique which squashes the high-intensity faulty activation values to alleviate their impact. We achieve this by replacing the unbounded activation functions with their clipped versions. We also present a method to systematically define the clipping values of the activation functions that result in increased resilience of the networks against faults. We evaluate our technique on the AlexNet and the VGG-16 DNNs trained for the CIFAR-10 dataset. The experimental results show that our mitigation technique significantly improves the resilience of the DNNs to faults. For example, the proposed technique offers on average 68.92\% improvement in the classification accuracy of resilience-optimized VGG-16 model at \num{1e-5} fault rate, when compared to the base network without any fault mitigation.

\end{abstract}


\section{Introduction}

\begin{figure}[t]
    \centering
        \captionsetup{justification=centering}
    \includegraphics[width=0.5\textwidth]{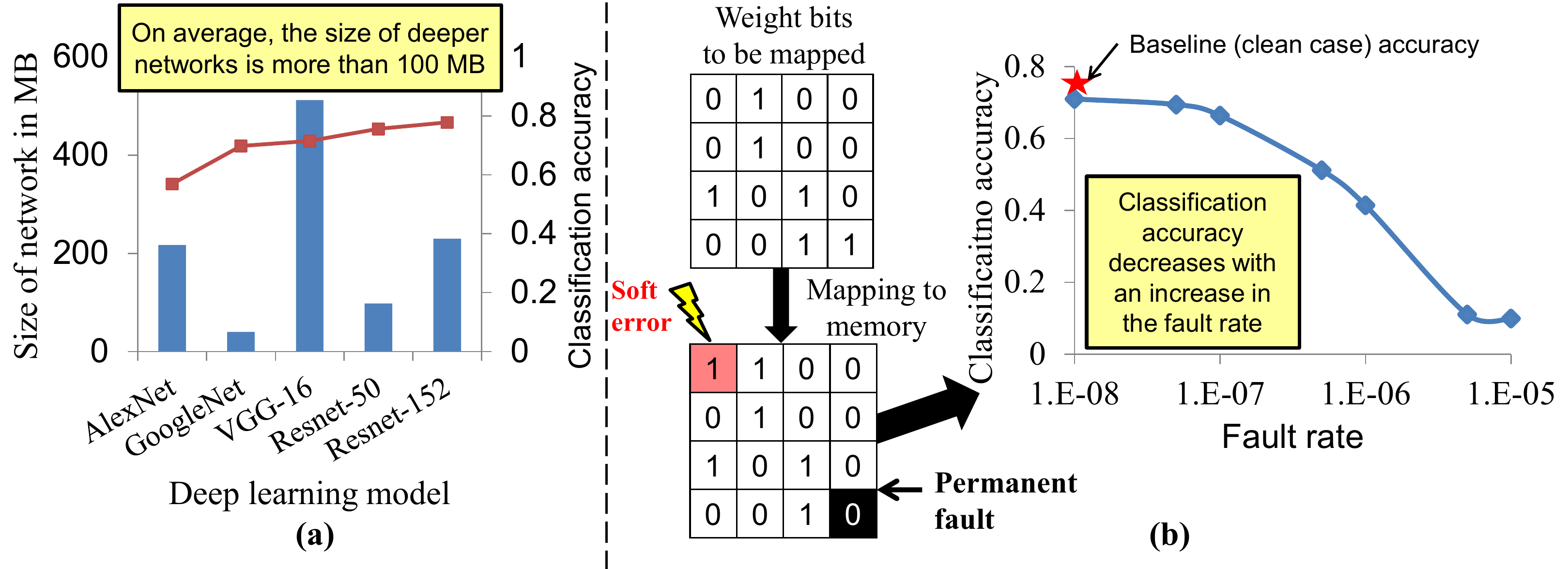}
    \caption{(a) Memory consumption of state-of-the-art DNN models. (b) The impact of hardware faults (bit flips in the weight memory) on the classification accuracy of AlexNet.}
    \label{fig:motivation}
\end{figure}

Due to their state-of-the-art accuracy in various applications, Deep Neural Networks (DNNs) have become the primary choice for most of the machine learning-based applications ~\cite{lecun2015deep}, ranging from simpler ones like hand written digit recognition to complex safety-critical applications like autonomous driving. In general, DNNs require a significantly large number of parameters (as shown in Fig.~\ref{fig:motivation}a for prominent DNNs used for image classification) to generalize well for real-time scenarios and, therefore, are highly computation and memory intensive. 
To efficiently process data using these networks, specialized hardware accelerators are utilized which are built using smaller technology nodes, in order to achieve high power and performance efficiency\cite{tpu},\cite{eyeriss},\cite{EIE}. Moreover, these accelerators make use of large on-chip and off-chip memories to store the parameters of the DNNs. 

A major concern that DNN accelerators face in the nano-scale technologies is their reliability against faults, i.e, they suffer from faults due to soft errors, aging and manufacturing-induced defects~\cite{vlsi}, which can lead to catastrophic effects in case of their usage in safety-critical applications~\cite{ISO}. Fig. \ref{fig:motivation}b illustrates our reliability analysis for the baseline AlexNet DNN (i.e. unprotected)~\cite{alexnet} doing image classification on the CIFAR-10 dataset~\cite{cifar}. It can be noticed that the accuracy drops significantly with growing error rates. 

Anecdotally, researchers speculated that DNNs forgive hardware errors~\cite{zhang2015approxann}. But, our analysis (and other studies like \cite{dac19_garg}) has revealed that the accuracy drops even at low/nominal fault rates. In this paper, through a comprehensive analysis, we will show that it highly depends upon which weights are corrupted and if they belong to the sensitive neurons or not. 
% An example of the impact of memory-faults on the classification accuracy of the modified AlexNet model\cite{alexnet} on the CIFAR-10 dataset\cite{cifar} is shown in Fig.~\ref{fig:motivation}b. 
% ==========>>>>>>>> OVER HERE.<<<<<<<<<<==================
% Due to the large storage requirement of weights and intermediate states in on- and off-chip memory during the execution of DNN, faults in these components result in mutated values which can lead to drastic effect on the DNNs. An example of the impact of memory-faults on the classification accuracy of the modified AlexNet model\cite{alexnet} on CIFAR-10 dataset\cite{cifar} is shown in Fig.~\ref{fig:motivation}b. In this work, we focus on the hardware faults in memory. 
\textit{In short, there is a dire need for improving the resilience of these networks to provide reliable functionality when used with unreliable hardware having nominal fault rates.}
% Deep Neural Networks have been widely adopted in various applications ranging from simple image classification to safety-critical applications like autonomous driving~\cite{lecun2015deep}. Normally, the size of DNN models will determine their classification accuracy. The higher their size, the better higher classification accuracy <this sentence should be revised>. Therefore a large number of parameters are required to be trained leading to significant amount of memory space to be used. Technology scaling offers the chance to have low-power and high-density memory chips. However, it also brings the challenge of increased fault rates which further results in unreliable hardware\cite{vlsi}. With the state-of-the-art networks being used in safety-critical applications, there is a significant need for making these networks reliable \cite{ISO}. 

\textbf{State-of-the-art and their Limitations:} Various techniques have been proposed to mitigate the effects of hardware-level faults in DNN-based systems. \textit{At hardware-level}, redundancy-based fault-mitigation techniques are commonly used, e.g., Dual Modular Redundancy (DMR) and Triple Modular Redundancy (TMR) \cite{TMR} for mitigating faults in computational units, and Error Correction Codes (ECC)\cite{ECC} for error-detection and correction in memories. In fact, the machine learning hardware in Tesla’s self-driving cars uses expensive DMR to mitigate the impact of faults\cite{tesla}. Note that, although these approaches offer improved resilience against faults, they have high overheads and are not preferable for computation/memory intensive DNNs. Other techniques include selective node hardening to improve the reliability of standard logic cells \cite{rel-synthesis} and hardened SRAM-cells \cite{tolerating,hardened-mem}. \textit{At software-level}, fault-aware training has been introduced for mitigating the memory faults~\cite{MATIC,RRAM}. However, there are two drawbacks of these approaches: (1) they require access to the training dataset, which in several real-world scenarios may not be available for designing Inference Engines~\footnote{For example, consider a DNN IP provided by a service provider, which has to be deployed on a particular embedded hardware. The IP provider has not made the training dataset available (as training dataset is a key IP), and one of the system requirements is to have a defined-level of fault-tolerance which the network (when deployed on the embedded hardware) does not meet.}; and (2) retraining costs a lot of resources and it may not be feasible to do it for every single chip. Moreover, such solutions are only limited to design-time faults, and cannot cope with run-time faults.

%\begin{itemize}
%    \item Using redundancy modules (e.g., Tripple Modular Redundancy \cite{TMR}) is one of generic mitigation techniques to make DNNs model resilient against hardware faults. However, this method comes with the high overhead due to the computational expensive of DNN models.
%    \item Error Correcting Code (ECC) using Hamming code ~\cite{ECC} can be alleviated to improve the error resilience of DNNs against hardware faults at memory. However, it comes with timing, area, or power cost. 
%    \item Fault-aware training is one of the commonly used method for improving the resilience of DNNs against hardware faults which demonstrates well in ~\cite{MATIC,RRAM}. Nevertheless, the drawback of this approach is that the full access to the training set is required, which is not available in some cases.
%    \item Hardened SRAM technique is used to tolerate soft errors in DNN accelerators~\cite{tolerating}. The drawback of this technique is its less flexibility to be deployed in different hardware platform. deep learning accelerator.
%\end{itemize}

\textbf{Targeted Research Problem:} How to improve the resilience of the DNNs to hardware-level faults with minimal energy/power and performance overhead and without the need of training dataset, redundancy, or any costly reliability feature.

%\begin{itemize}
%   \item If DNN models running on resource-constrained hardware platform like mobile devices, using redundancy modules or ECC technique is not an optimal option. It poses a question if we can have a low-cost mitigation technique that working in the above-mentioned platform while maintaining the resilience of the DNN models.
%    \item In some cases, fault-aware training is not feasible to deploy as training dataset is not available. Let's consider the scenario: we are given pre-trained DNN models and we can generate a small subset of validation dataset. The question is whether we can design a mitigation technique benefit from this small subset to make the DNN models robust against the hardware faults.
%    \item If given DNN models are optimized and any modification of the parameters during retraining may lead to the degradation of the DNN models, e.g. the drop of classification accuracy of DNN models for image classification task, fault-aware training is not a feasible option. Therefore, it raises the question of whether we can use a mitigation technique which does not modify the optimized parameters of the given DNN models while making them resilient.
%\end{itemize}

\textbf{Our Novel Contributions:} We address the above challenge through the following novel contributions:
\begin{itemize}
    \item We perform a comprehensive analysis (Section~\ref{sec:analysis}) to study the impacts of hardware-level faults on the accuracy and the intermediate outputs of the DNNs. This allows us to understand the resilience of DNNs in a systematic way, which can enable an efficient reliability mechanism. 
    \item Based on the analysis, we propose a clipped activation function (Section~\ref{sec:mitigation_technique}) for improving the resilience of the DNNs, which bounds the intermediate output (i.e., activation) values of the networks to a defined range. 
    \item We propose a systematic methodology (Section~\ref{sec:mitigation_technique}) to define the output range of the activation functions for each layer of a DNN without the need of the training dataset and without modifying the weights and biases of the network.
    \item We present a comprehensive evaluation of the effectiveness of our mitigation technique on the AlexNet and the VGG-16 networks. The evaluation shows 18.19\% and 69.49\% improvement in the classification accuracy of the AlexNet and the VGG-16 networks, respectively, at $5 \times 10^{-7}$ fault rate compared to their baseline (without error mitigation) variants.
    % \item An open-source fault injection framework will be publicized after the publication
    % \item We perform comprehensive evaluations for the AlexNet and the VGG-16 network trained on the CIFAR-10 dataset with fault injection. The results illustrate that the proposed method provides - and - improvements in the classification accuracy of the AlexNet and the VGG-16 
    %\item Using  In this sense, we replace the unbounded activation function with a novel clipping activation function (i.e., bounded activation function) which squashes the activation to a defined range.
    %//In this sense, we present a novel clipping activation function which squashes the activation to a defined range. We only replace the unbounded activation function with the bounded one==>>> |||| ||||
    % Toward this, our mitigation technique requires less overhead of the resource compared with ECC technique or redundancy modules.
    %\item A methodology to select the threshold of the clipping activation function for each layer% by fine-tuning the threshold.
    %\item Our mitigation technique does not modify parameters of the optimized DNN models.
\end{itemize}

% \textbf{Paper organization:}  Section II provides a brief overview of DNNs. In Section III, we presents the comprehensive analysis to study the impact of hardware-level faults on the accuracy and the intermediate outputs of the DNNs. Section IV describes our systematic methodology to mitigate the effects of hardware-level faults without the need of re-training. In Section V, we present the results and discussion. At the end, Section VI concludes the paper.

\section{Background: An Overview of DNNs}
\begin{figure}[t]
    \centering
    \captionsetup{justification=centering}
    \includegraphics[width=0.9\linewidth]{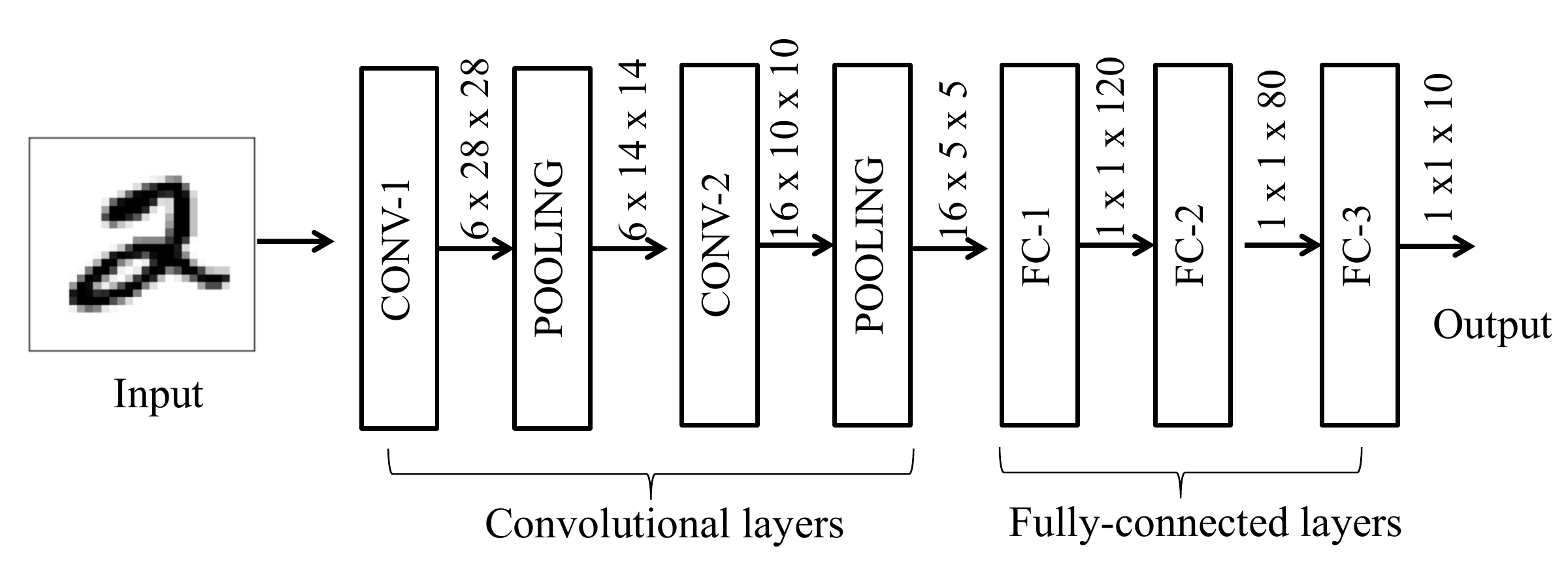}
    \caption{A high-level view of the LeNet-5 network}
    \label{fig:background_alexnet}
\end{figure}

A prominent type of DNNs is Convolutional Neural Networks (CNNs), which is used for processing spatially correlated data, e.g., images and videos. A CNN is mainly composed of two types of computational layers, i.e., convolutional (CONV) layers and fully-connected (FC) layers, where each computational layer is followed by an activation layer and each CONV layer is (optionally) followed by a pooling layer. 
Note that the FC layers are used for classification tasks and, therefore, are used towards the end of the CNNs while the CONV layers are used for extracting features and, therefore, are placed at the start and feeds the extracted features to the FC layers. A high-level view of the LeNet-5 network is shown in Fig.~\ref{fig:background_alexnet}. The outputs of these layers are generated by the dot product operations between parameters and input values, which are then passed through activation functions, e.g. ReLU, to add non-linearity in the computations. The outputs from the activation functions are usually referred to as activations. A more comprehensive overview of the neural networks can be found in~\cite{sze2017efficient}. 
% Equation <Le-Ha TODO>
% The lifetime of a DNN can be divided into two phases, i.e., training phase and inference phase. In the training phase, the network parameters are tuned that it provides high accuracy for a given task. This is performed using a back-propagation algorithm and a training dataset. Once trained, the model is then deployed in systems to process data, which is referred to as inference phase. In this paper, we focus only on the inference phase. A more comprehensive overview of the neural networks can be found in~\cite{sze2017efficient}. 
% The deployment of CNNs consists of two phases: training phase and inference phase. The training phase is performed once in which parameters of the model are tweaked to achieve the expected accuracy. Then the model can be deployed into a platform to perform the inference phase. Current state-of-the-art DNN models are complex which stores a large number of parameters to maintain the high accuracy. A more comprehensive overview of the neural networks can be found in ~\cite{sze2017efficient}.

\section{Error Resilience Analysis of Deep Neural Networks}
\begin{figure*}[t]
    \centering
        \captionsetup{justification=centering}
    \includegraphics[width=0.95\textwidth]{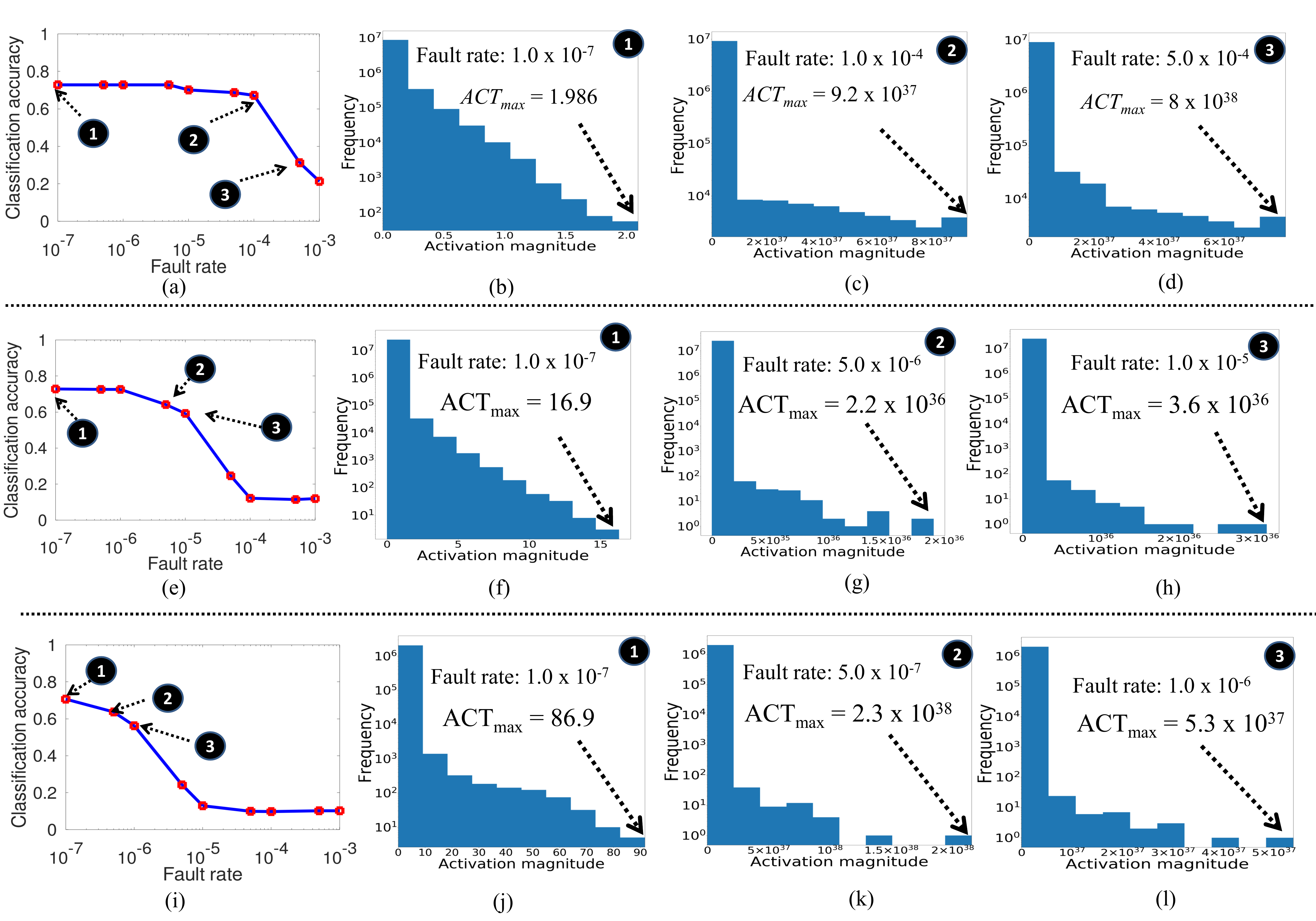}
    \caption{Error resilience analysis of CONV-1 layer (a-d), CONV-5 layer (e-h), and FC-1 layer (i-l) of the AlexNet on the CIFAR-10  dataset}
    \label{fig:error_analysis}
\end{figure*}

\label{sec:analysis}

To analyze the error resilience of a DNN against memory faults, we developed a fault-injection framework, where random bit-flips are injected in the memory blocks storing the parameters of the DNN model. 
We perform per-layer fault injection to study the sensitivity of individual layers and the effects of the faults on the output activations. 
Fig.~\ref{fig:error_analysis} illustrates the resilience of CONV-1 layer (first computational layer), CONV-5 layer (fifth computational layer), and FC-1 layer (sixth computational layer) of the AlexNet. The figure also shows the distributions of the output activations of the respective layers. 

From the analysis of Fig.~\ref{fig:error_analysis}, we draw the following key observations:
\begin{itemize}
    \item In general, the classification accuracy of the network decreases with an increase in the fault rate, as shown in Figs.~\ref{fig:error_analysis}a,~\ref{fig:error_analysis}e, and~\ref{fig:error_analysis}i.  Moreover, the decrease in the accuracy is monotonic, which is mainly because, at higher fault rates, the probability of a fault occurring at a critical location is significantly higher. 
    \item At lower fault rates, the accuracy of the network stays close to the baseline accuracy before dropping drastically, as there is a significant chance that the faults do not occur at critical bit locations or are masked within the network. Also, the fault rate till which the accuracy stays close to the baseline accuracy is different for each layer. This is because each layer has different number of parameters and has different number of layers between the output and itself. 
    \item The distribution of the output activations at higher fault rates have values of higher-intensities as well, as can be observed from Figs.~\ref{fig:error_analysis}c and~\ref{fig:error_analysis}d. This trend is consistent across layers, as can also be seen in Figs.~\ref{fig:error_analysis}g,~\ref{fig:error_analysis}h,~\ref{fig:error_analysis}k, and~\ref{fig:error_analysis}m. This is mainly because of the fact that the weights are distributed close to zero value and bit-flips from 0 to 1 at Most Significant Bit (MSB) locations of the weights can result in them having higher magnitudes and, thereby, resulting in high-intensity activations during inference. 
    %\item We perform a comprehensive resilience analysis of DNN models by launching fault njection to weights of the pre-trained modified AlexNet model for image classification task on CIFAR-10 dataset\cite{cifar}. Toward this, we investigate per-layer sensitivity of the network by flipping the bits of the weights randomly. The mutated model is then evaluated on a subset of CIFAR-10 validation dataset. We also profile the maximum activation ($ACT_{max}$) and activation histograms at designated fault rates and fault patterns, at various fault rates. For example, we demonstrate the analysis on the first layer (CONV-1 layer), fifth layer (CONV-5 layer) and sixth layer (FC-1 layer) in Fig. \ref{fig:conv1_analysis}, Fig. \ref{fig:conv5_analysis}, and Fig. \ref{fig:fc1_analysis}.
    % \item One key observation drawn from this analysis is that there are existing \textbf{\textit{high-intensity activations}}at higher fault rate. In case Most Significant Bits of the weights of the layer are flipped from O to 1 due to hardware faults, these parameters are mutated to high-intensity values resulting in high-intensity activation. Consequently, the classification accuracy drops drastically.
\end{itemize}

\section{Our Mitigation Technique for Improving Fault Tolerance of Deep Neural Networks }
\label{sec:mitigation_technique}
\begin{figure}[t]
    \centering
    \captionsetup{justification=centering}
    \includegraphics[width=0.5\textwidth]{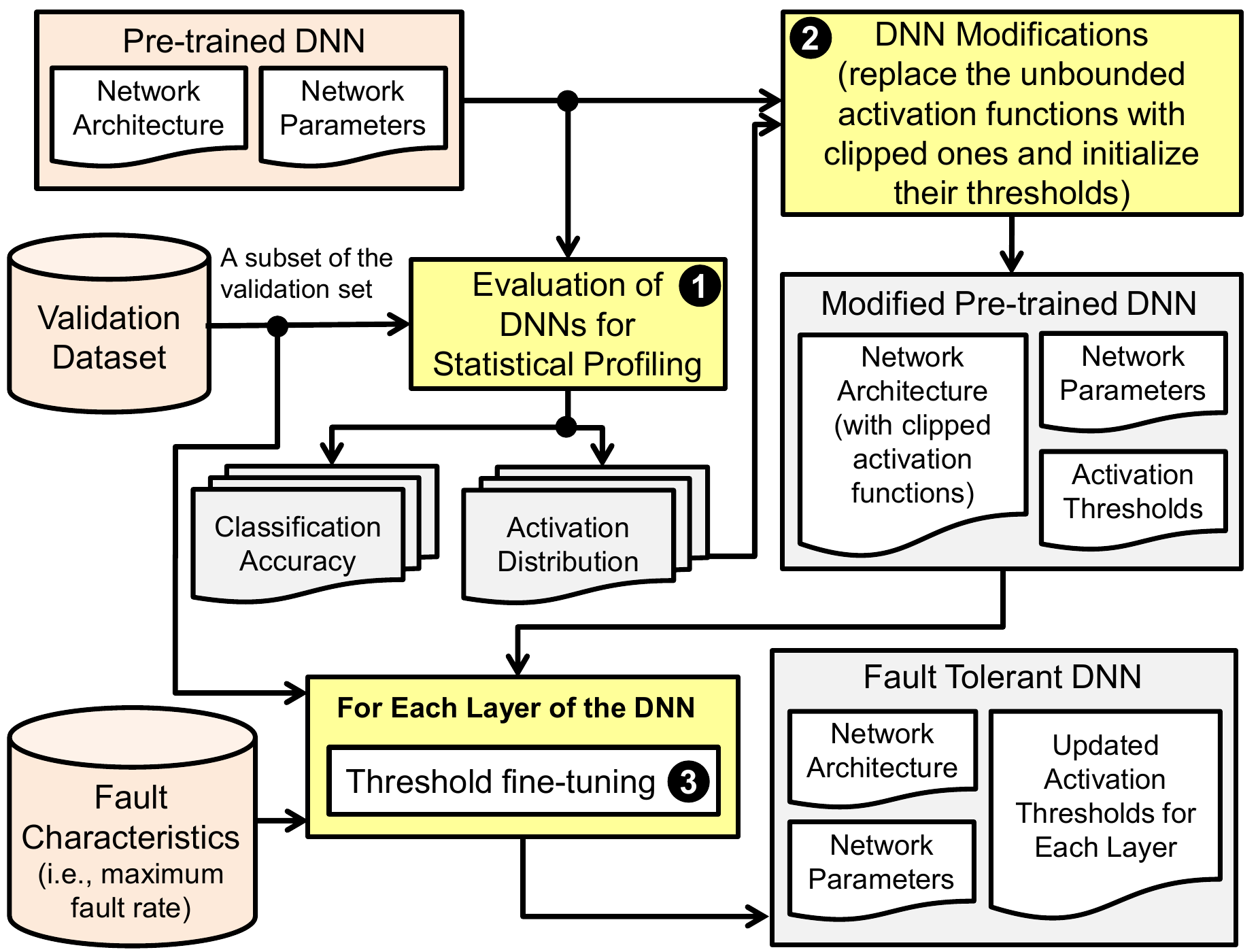}
    \caption{Our methodology to improve the resilience of a pre-trained DNN model}
    \label{fig:methodology}
    \vspace{-3mm}
\end{figure}

Fig.~\ref{fig:methodology} shows an overview of our methodology for improving the fault tolerance of DNNs using \textit{clipped activation functions}. The methodology is based on the observation made in Section~\ref{sec:analysis} that higher fault rates result in faulty activations with higher magnitudes, which dominate the result and may lead to misclassification. The proposed methodology is independent of the training dataset and only requires a small subset of the validation set for tuning the clipping thresholds of the clipped activation functions. 
Our methodology operates in three main steps, as discussed below. 

\textbf{Step-1:} We perform profiling for computing the statistical properties of the activations of all the layers using a subset of the validation dataset. The statistics extracted from this step are the maximum value of the activations ($ACT_{max}$) observed at the output of each layer. 
% The statistics extracted from this step is the set of maximum activation value ($ACT_{max}$) observed at the output of each layer. 

\textbf{Step-2:} We  replace  the  unbounded  activation  functions  in  the  DNN with  their  clipped  variants  (explained  in  Section~\ref{sec:activation_function})  and initialize  their  thresholds  with  their  corresponding $ACT_{max}$. 

\textbf{Step-3:} We perform fine-tuning of the clipping thresholds using an efficient method explained in Section~\ref{sec:Threshold_Fine-Tuning}. The metric used for resilience evaluation is presented in Section~\ref{sec:threshold-accuracy_relation}. Note that Step~3 is repeated for each layer of the network, using the network generated from Step~2, to find suitable clipping thresholds for all the layers. \textit{The final outcome from the methodology is a fault-tolerant DNN with optimized thresholds for the clipped activation functions.}

\subsection{The Clipped Activation Function}
\label{sec:activation_function}
\begin{figure}[b]
\vspace{-3mm}
    \centering
        \captionsetup{justification=centering}
    
    \includegraphics[height=0.22\textwidth, width=0.5\textwidth]{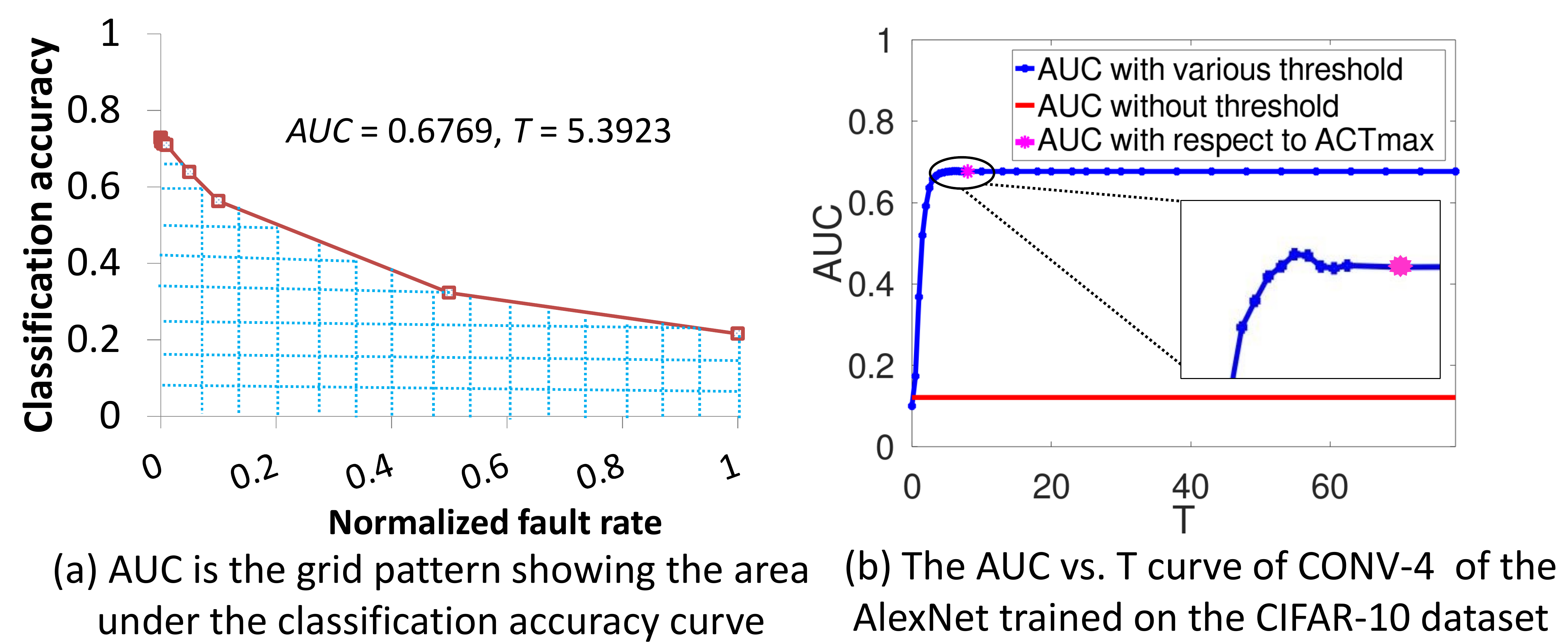}

    \caption{(a) Illustration of the $AUC$ calculation of a the AlexNet with clipped activation in the presence of faults in CONV-4 layer; (b) The AUC curve with various threshold (blue line) and AUC without threshold (red line)}
    \label{fig: AUC show}
\end{figure}
Based on the observations made in Section~\ref{sec:analysis} and following an inspiration from the pruning~\cite{han2015deep} and the dropout~\cite{srivastava2014dropout} techniques, we introduce a novel \textit{clipped version of the ReLU activation function} for mapping high-intensity (possibly faulty) activation values to zero. We formulate this function as:
\vspace{-0.5mm}
\begin{gather*}
f(x) = \begin{cases}
x,\ if\ 0\ \leq x \leq T \\
0,\ otherwise\\
\end{cases}   
\end{gather*}
\vspace{-2.5mm}

Where, $f(x)$ is the output activation, $x$ is the input (i.e., output after dot-product operation), and $T$ is the clipping threshold beyond which all the values are considered faulty and are mapped to zero. Although we present the clipped version of only the ReLU function, clipped versions of other activation functions (e.g., Leaky-ReLU) can also be designed similarly. 

\subsection{Resilience Evaluation Metric and the Corresponding Analysis for Finding Suitable Clipping Thresholds}
\label{sec:threshold-accuracy_relation}
\begin{figure}[b]
\vspace{-3mm}
    \centering
        \captionsetup{justification=centering}
        \includegraphics[width=1\linewidth]{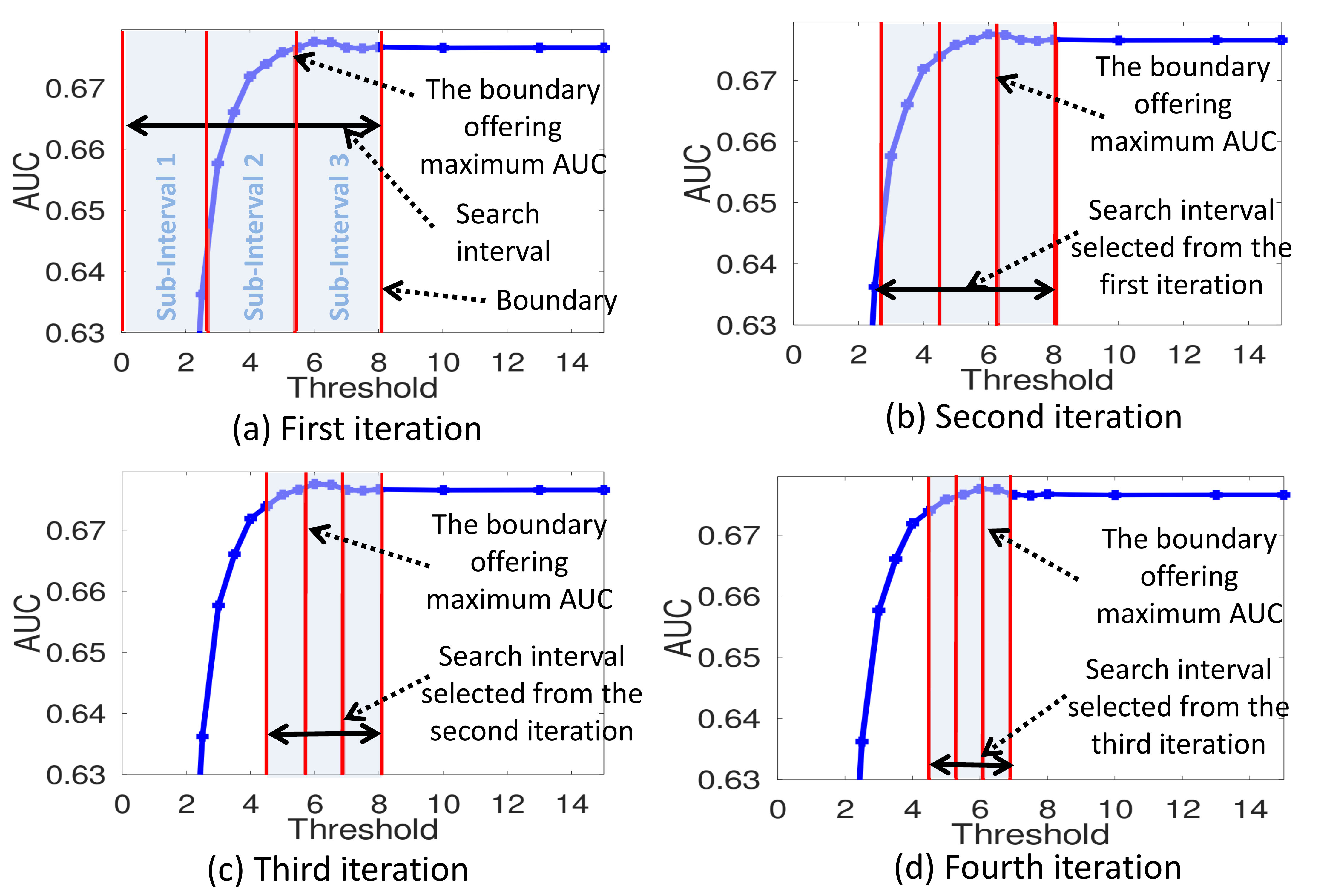}
        
    \caption{Threshold Fine-Tuning Algorithm applied on CONV-4 layer of the AlexNet on CIFAR-10 dataset}
    \label{algorithm illustration}
\end{figure}

\textbf{Evaluation metric:} Hardware fault-rates can vary in a defined range in real scenarios. Therefore, to capture the resilience characteristics of a network across different fault rates in a single metric, we introduce the area under the accuracy vs. normalized fault rate curve ($AUC$) as a metric, where the area is computed using the Trapezoidal rule. 
% for a defined set of network parameters, we introduce the area under the accuracy vs. normalized fault rate curve ($AUC$) as a metric, where the area is computed using the Trapezoidal rule. 
An illustration of this is shown in Fig.~\ref{fig: AUC show}a, where the area of the region marked with blue grid represents the $AUC$. Note that both the axes are normalized such that the ideal scenario, i.e., the case where the network provides 100\% accuracy at all the considered fault rates, has an $AUC$ of 1. 
%\begin{itemize}
%    \item $AUC$ is the area under the classification curve drawn from discrete pairs of accuracy mean and normalized fault rate.
%    \item For example, in Fig. \ref{fig: AUC show}a, $AUC$ is the area of the grid pattern shade under classification accuracy curve when faults are injected into CONNV-4 layer using clipping activation with $T$ = 5.3923.
%\end{itemize}

\textbf{Resilience sweep across thresholds:} 
To study the impact of threshold value of the clipped activation function of a layer of a network on the resilience of the network, let us consider the $AUC$ vs. $T$ curve of CONV-4 layer of the AlexNet network trained on the CIFAR-10 dataset. The plot is shown in Fig.~\ref{fig: AUC show}b. As can be seen from the figure, moving from higher to lower threshold values, the $AUC$ rises to a peak value at a particular location before decreasing drastically. Ideally, we should select the threshold at this location as the optimal clipping threshold ($T_{optimum}$) of the activation functions of the layer, as it offers the highest resilience within a pre-defined fault range. Note that, although the blue curve in the figure seems to have a fixed value at higher $T$ values (i.e., at $T$ > $ACT_{max}$), the $AUC$ of the network with unbounded activations is significantly low, as shown with the help of red line in the figure. This also reaffirms the fact that clipping high-intensity activation values can significantly improve the overall resilience of a DNN. 

\subsection{Threshold Fine-Tuning Algorithm}
\label{sec:Threshold_Fine-Tuning}

% \begin{figure}[t]
% \begin{algorithm}[H]
\begin{algorithm}

%%%%
\caption{Threshold Fine-Tunning  }
\label{alg:the_alg}
\footnotesize

  \SetKwData{Left}{left}
  \SetKwData{Up}{up}
  \SetKwFunction{FindCompress}{FindCompress}
  \SetKwInOut{Input}{input}
  \SetKwInOut{Output}{output}
  \Input{Modified pre-trained DNN model from Step 2 of the proposed methodology}
  \Output{\textit{T}}

%%%%%%%%%%%%%%%%%%%%%%%%%%%%%%%%%%%%%%55

\textbf{BEGIN ALGORITHM}\\ 
\textit{counter} $\leftarrow$ 1 ; \\
\While{counter $\leq$ N}
{

    \If{i ==1}{
        S $\leftarrow$ [0 ,\ $ACT_{max}$] ;
        
        $AUC_{1}^{T_{1}}$, $AUC_{2}^{T_{2}}$, $AUC_{3}^{T_{3}}$, $AUC_{4}^{T_{4}}$ = $AUC\_Calculation(S)$ ;\\
        }
    \Else{
        S,\ T = Interval\_Search( $T_{1}$, $T_{2}$, $T_{3}$, $T_{4}$, $AUC_{1}^{T_{1}}$, $AUC_{2}^{T_{2}}$, $AUC_{3}^{T_{3}}$, $AUC_{4}^{T_{4}}$) ;
        
        $AUC_{1}^{T_{1}}$, $AUC_{2}^{T_{2}}$, $AUC_{3}^{T_{3}}$, $AUC_{4}^{T_{4}}$ = AUC\_Calculation(S) ;\\
        }
    counter $\leftarrow$ counter+1 ;\\   
\For{i = 1 to 3}
    { $\Delta_{i}$ $\leftarrow$ $\mid$ $AUC_{i+1}^{T_{i+1}}$ - $AUC_{i}^{T_{i}}$ $\mid$; 
    }
\If{ maximum($\Delta_{1}$, $\Delta_{2}$, $\Delta_{3}$)  $\leq$ $\delta$ and counter $\geq$ M}
    {
    \textbf{Return} T ;
    }
}  
\textbf{Return} T ;   

\textbf{END ALGORITHM} 

\SetKwFunction{FMain}{Interval\_Search}
\SetKwProg{Fn}{Function}{:}{}
\Fn{\FMain{$T_{1}$, $T_{2}$, $T_{3}$, $T_{4}$, $AUC_{1}^{T_{1}}$, $AUC_{2}^{T_{2}}$, $AUC_{3}^{T_{3}}$, $AUC_{4}^{T_{4}}$}}
{
Index $\leftarrow$ index\ of\ T\ with\ the\ highest\ AUC\\
 \If{Index == 4}{
    $\Bar{S}$ $\leftarrow$ [$T_{3}$,\ $T_{4}$] \;
  }
  \uElseIf{Index == 1}{
    $\Bar{S}$ $\leftarrow$ [$T_{1}$ ,\ $T_{2}$] \;
                        }
  \Else {
   $\Bar{S}$ $\leftarrow$ [$T_{Index-1}$,\ $T_{Index+1}$] \;
        }
  T $\leftarrow$ $T_{Index}$ ;\\
\textbf{Return} $\Bar{S}$, T ;\\
}

\SetKwFunction{FMain}{AUC\_Calculation}
\SetKwProg{Fn}{Function}{:}{}
\Fn{\FMain{S}}
{
$T_{1}$ $\leftarrow$ minimum(S) ;

$T_{2}$ $\leftarrow$ \ $T_{1}$ + (maximum(S) - minimum(S))/3 ;

$T_{3}$ $\leftarrow$ \ $T_{2}$ + (maximum(S) - minimum(S))/3 ;

$T_{4}$ $\leftarrow$ maximum(S); 

\For{i = 1 to 4}
    {
    Evaluating model using $T_{i}$;
    
    Calculating $AUC_{i}^{T_{i}}$;
    }
\textbf{Return} $AUC_{1}^{T_{1}}$, $AUC_{2}^{T_{2}}$, $AUC_{3}^{T_{3}}$, $AUC_{4}^{T_{4}}$;
}

% \end{algorithm}
\end{algorithm}
% \end{figure}

%%%%%%%%%%%%%%%%%%%%%%%%%%%%%%%%%%%%%%55

% The \textbf{challenge}\textit{} is how to define the $T$ for each layer. There is a need to have a profiling of the impact of $T$ on the error resilience of the DNNs in the presence of faults. For example, a small $T$ leads to the large pruned connections and may lead to the degradation of the accuracy. Whereas, a large $T$ may lead to inefficient mitigation technique as it cannot squash high-intensity activations. Therefore, there is a need to extract the Accuracy-Threshold relation. To this end, We introduce a novel metric which can be used to extract this relation efficiently. 

The threshold fine-tuning algorithm is based on the observation made in the previous subsection that the $AUC$ vs. $T$ curves always have a bell shaped curve, as also shown in Fig.~\ref{fig: AUC show}b. Another key observation which helped us in designing an efficient algorithm is that the peak of the curve always lie below the $ACT_{max}$ value determined in Step~1 of the methodology. 
The algorithm starts by initializing search interval, i.e., $S=[0,\ ACT_{max}]$, and dividing it into three equally-sized sub-intervals, which is illustrated in Fig.~\ref{algorithm illustration}a. 
The $AUC_{i}^{T_{i}}$ corresponding to the $i^{th}$ boundary, at the threshold $T_{i}$ in the current search interval, is computed for each $ i \in \{1,2,3,4\}$. The region ($\Bar{S}$) covering the sub-interval/s around the boundary offering maximum $AUC$ is selected while the rest are discarded. 
The search interval $S$ is updated with $\Bar{S}$ and then again divided into three equally spaced sub-intervals in the next iteration and the same process is repeated, as shown in Figs.~\ref{algorithm illustration}b,~\ref{algorithm illustration}c, and~\ref{algorithm illustration}d. 
This process is applied until the number of iterations ($counter$) reaches a defined number ($N$), or the maximum difference between the adjacent $AUC_{i}^{T_{i}}$s ($\Delta_{j}$, 1 $\leq$ j$\leq$ 3) is less than a predefined limit ($\delta$) and $counter$ $\geq$ $M$ ($M$ < $N$).
%The $AUC_{i}^{T_{i}}$ corresponding to the threshold $T_{i}$ ($\forall$ 1 $\leq$ i$\leq$ 4) at the boundaries of the sub-intervals is computed and the region ($\Bar{S}$) covering the sub-intervals around the boundary offering maximum $AUC$ are selected while the rest are discarded. 
%The search interval $S$ is updated with $\Bar{S}$ and then again divided into three equally spaced sub-intervals and the same process is repeated for the next iterations, as shown in Figs.~\ref{algorithm illustration}b,~\ref{algorithm illustration}c, and~\ref{algorithm illustration}d. 
%This process is applied until number of iterations ($counter$) reaches a defined number ($N$), or the maximum difference ($\Delta_{i}$, 1 $\leq$ i$\leq$ 3) between the $AUC_{i}^{T_{i}}$ is less than a predefined limit ($\delta$) and $counter$ $\geq$ $N$ ($N$ < $M$). 
The detailed algorithm is shown in Algo.~\ref{alg:the_alg}.

\section{Results And Discussion}
\label{sec:results_and_discussion}
\subsection{Experimental setup}
We evaluated our proposed mitigation technique on two DNNs models, i.e., the AlexNet and the VGG-16\cite{vgg-16}. Both the models are modified to take the CIFAR-10 dataset images as inputs. The AlexNet contains 5 CONV layer and 3 FC layer while the base VGG-16 contains 13 CONV layer and 1 FC layer. The AlexNet and the VGG-16 offer baseline classification accuracies of 72.8\% and 82.8\%, respectively. % The CIFAR-10 test dataset is used at the last stage of experiment to evaluate error resilience of networks using our proposed mitigation technique in the sense of faults are randomly injected across the networks.
% What is the range of the fault rate that you have assumed? From 1E-8 to 5E-5 for VGG-16; From 1E-8 to 1E-5 for AlexNet

We developed our fault injection framework in Python using the Pytorch framework\cite{pytorch}. The developed framework is in-line with other fault injection frameworks proposed in state-of-the-art works, e.g., Ares in~\cite{Reagen}. All experiments are performed on an Intel Core i7@3.2 GHz processor with two NVIDIA GeForce GTX 1080 Ti GPUs. 

\subsection{Comparison with the unprotected DNNs}

\begin{figure*}[t]
    \centering
    \captionsetup{justification=centering}
    \includegraphics[width=0.92\textwidth]{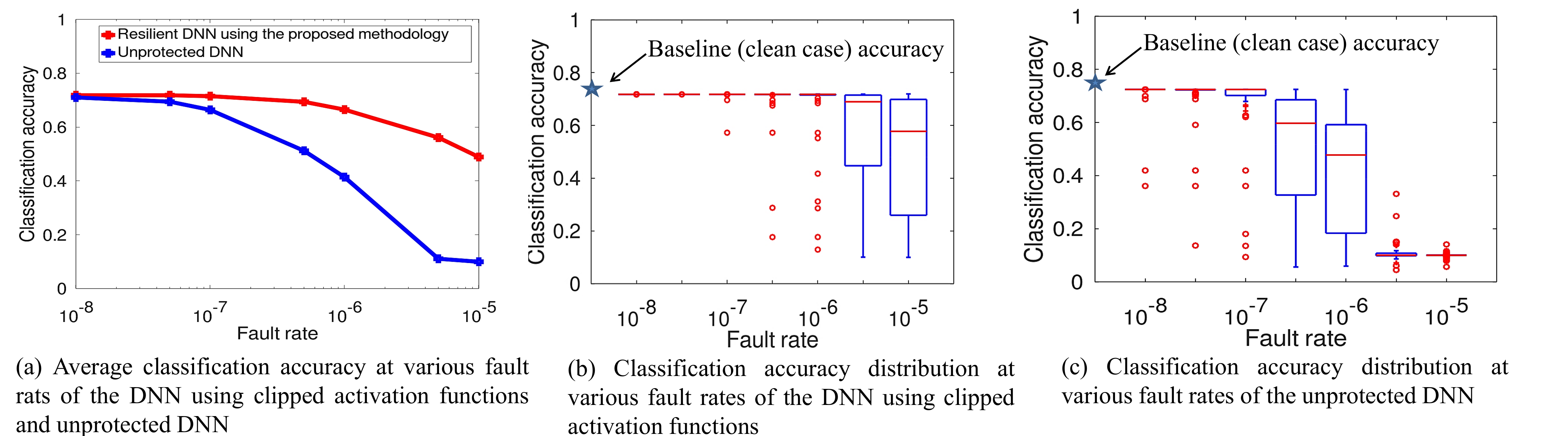}
    \vspace{-1mm}
    \caption{Error resilience evaluation of the AlexNet with and without clipped activation functions}
    \label{fig: AlexNet}
\end{figure*}

To show the effectiveness of the proposed methodology, we compared the accuracy of the resilient DNNs, developed using the proposed method, with unprotected DNNs. Fig.~\ref{fig: AlexNet}a shows the classification accuracies of the resilient and the unprotected AlexNet. The figure clearly illustrates that the network with clipped activation functions shows significant improvements in the fault-resilience of the DNN at fault rates around \num{1e-7} and $1 \times 10^{-6}$. For example, the classification accuracy of the resilient AlexNet with clipped activations at $5 \times 10^{-7}$ fault rate is 69.36\% compared to 51.16\% observed for the unprotected DNN. Overall, the proposed method shows 173.32\% improvement in the $AUC$ of the AlexNet considering the fault range from 0 to $1 \times 10^{-5}$. Note that the accuracies reported in Fig.~\ref{fig: AlexNet}a are mean values computed using 50 experiments, which is already large considering highly compute-intensive nature of DNNs and their multiple execution runs and parameter settings. 

Figs.~\ref{fig: AlexNet}b and~\ref{fig: AlexNet}c show the variations across multiple experiments using box plot. Note that at fault rates $1 \times 10^{-8}$ and $5 \times 10^{-8}$ the worst-case accuracy of the resilient network, generated using the proposed methodology, is close to the baseline accuracy (i.e., 72.8\%) while the worst-case accuracy of the unprotected network for the same fault rates is 41.93\% and 13.66\%, respectively, i.e., significantly lower than the baseline. 

% Figs.~\ref{fig: AlexNet}b and~\ref{fig: AlexNet}c show the variations across multiple experiments using box plot. It is noticed in the Figures~\ref{fig: AlexNet}b that at fault rates \num{5e-7} and \num{1e-6} the accuracy distributions are skewed to the baseline accuracy, i.e. 72.8\%, with only a few extreme cases. Whereas, the accuracy distributions encounter great variability in Figures~\ref{fig: AlexNet}c.

\begin{figure*}[h]
    \centering
    \captionsetup{justification=centering}
    \includegraphics[width=0.92\textwidth]{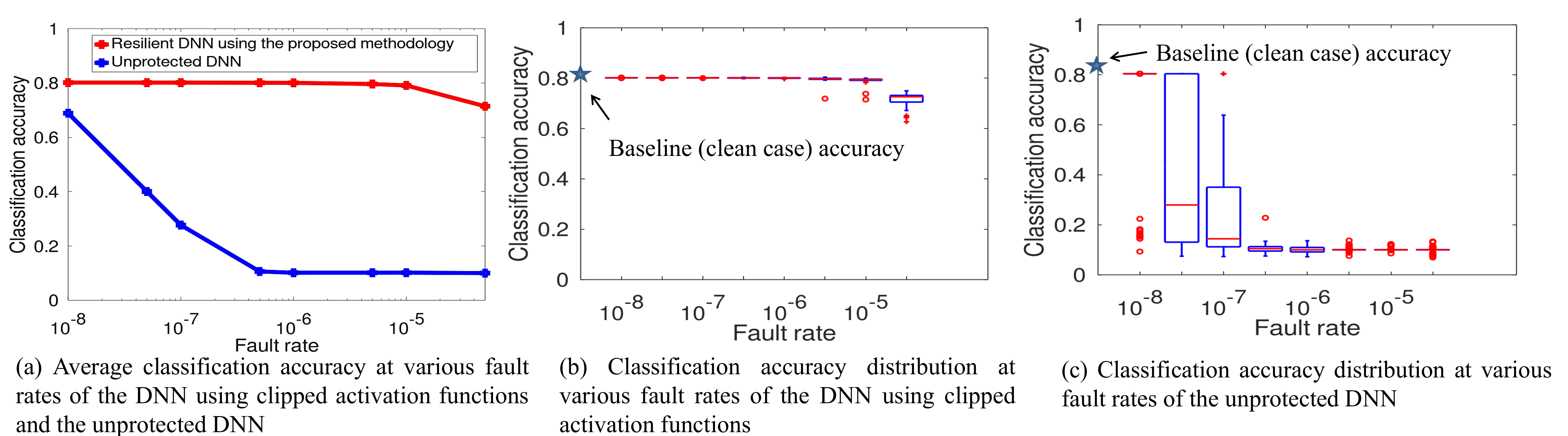}
    \vspace{-1mm}
    \caption{Error resilience evaluation of the VGG-16 with and without clipped activation functions}
    \label{fig: vgg16}
\end{figure*}

Similar trend is observed in case of the VGG-16 network, as shown in Fig.~\ref{fig: vgg16}. However, the proposed technique shows significant improvements in the resilience of the network, e.g., 654.91\% at fault rate \num{5e-7} in $AUC$ as can be observed from Fig.~\ref{fig: vgg16}a, even better than the case of the AlexNet network. 

% Similar trend is observed in case of the VGG-16 network, as shown in Fig.~\ref{fig: vgg16}. However, the proposed technique shows significant improvements in the resilience of the network. For example, the $AUC$ is improved significantly up to 654.91\% at fault rate \num{5e-7}. In detail, the accuracy of network using our proposed technique is 80.09\%, which is close to the baseline accuracy, compared with 10.15\% accuracy of the unprotected network.

\textit{Note that for all the results reported in Figs.~\ref{fig: AlexNet} and~\ref{fig: vgg16}, we employed the CIFAR-10 test set in order to avoid any overlap between the data used for testing and the data used for computing the thresholds.}

\section{conclusion}
In this work, we presented an analysis to study the impact of hardware faults on the accuracy and the intermediate outputs of the DNNs. We analyzed how high-intensity activations, generated due to the parameter corruption, result in the degradation of the accuracy of DNN models. To mitigate the effects of faults, we proposed a technique based on clipped activation functions, which blocks the high-intensity (potentially faulty) activations and maps them to zero. We also proposed an efficient algorithm for defining the range of the clipped activation functions. 
The proposed technique offers a significant improvement in the resilience of the DNNs. For example, the proposed technique provides 68.92\% improvement at $10^{-5}$ fault rate for the VGG-16 network trained on the CIFAR-10 dataset, when compared to the unprotected network. 

% In this work, we presented a comprehensive analysis to study the impact of hardware faults on the accuracy and the intermediate outputs of the DNNs. We analyzed how high-intensity activations, generated due to the parameter corruption, result in the degradation of the accuracy of DNN models. To mitigate the effects of faults, we proposed a technique based on clipped activation functions, which blocks the high-intensity (potentially faulty) activations and maps them to zero. We also proposed an efficient algorithm for defining the range of the clipped activation functions. 
% The proposed technique offers significant improvement in the resilience of the DNNs. For example, the proposed technique provides 68.92\% improvement at $10^{-5}$ fault rate for VGG-16 network, when compared to the unprotected network.


\section*{Acknowledgement}
%\footnotesize
This work was supported in parts by the German Research Foundation (DFG) as part of the priority program ``Dependable Embedded Systems" (SPP 1500-spp1500.itec.kit.edu)

\vspace{6pt}
%\clearpage
\bibliographystyle{IEEEtran}
% \bibliography{biblio}
\bibliography{FTClipAct}

\end{document}